\pdfoutput=1
\documentclass[11pt]{article}

\usepackage{emnlp2021}
\usepackage{times}
\usepackage{latexsym}
\usepackage[T1]{fontenc}
\usepackage[utf8]{inputenc}
\usepackage{microtype}
\usepackage{graphicx}
\usepackage{multirow}
\usepackage{xcolor}
\usepackage{amsmath,amssymb,stackengine}
\usepackage{array}

\usepackage{svg}

\author{%
Daniela Brook Weiss$^{\,1}$%
~\;~\;~ Paul Roit$^{\,1}$%
~\;~\;~ Ori Ernst$^{\,1}$%
{\bf~\;~\;~ Ido Dagan$^{1}$}\\
  $^{1}$Computer Science Department, Bar-Ilan University\\%
  \texttt{\footnotesize{\{dani.b.weiss,plroit,oriern\}@gmail.com}}\, \texttt{\footnotesize{dagan@cs.biu.ac.il}}
}

\title{
Extending Multi-Text Sentence Fusion Resources \\ via Pyramid Annotations}

\date{}

\begin{document}
\maketitle
\begin{abstract}
NLP models that compare or consolidate information across multiple documents often struggle when challenged with recognizing substantial information redundancies across the texts. For example, in multi-document summarization it is crucial to identify salient information across texts and then generate a non-redundant summary, while facing repeated and usually differently-phrased salient content. To facilitate researching such challenges, the sentence-level task of \textit{sentence fusion} was proposed, yet previous datasets for this task were very limited in their size and scope. In this paper, we revisit and substantially extend previous dataset creation efforts. With careful modifications, relabeling and employing complementing data sources, we were able to triple the size of a notable earlier dataset.
Moreover, we show that our extended version uses more representative texts for multi-document tasks and provides a larger and more diverse training set, which substantially improves model training.

\end{abstract}
\section{Introduction}
Despite recent advances in summarizing single documents, multi document summarization (MDS) has not progressed with the same pace. 
The task still poses the same challenges of its single-document counter-part, namely addressing saliency and coherency, but it also requires effective measures for identifying and consolidating redundant information.
In light of this, several works proposed a tighter-scoped task, called \emph{sentence fusion}, which
focuses on summarizing multiple sentences with overlapping content into a non-redundant one. 
Such a sentence-level task allows a fine-grained analysis of which information units are shared among the input sentences, as well as control over different degrees of information inclusion and exclusion. Sentence fusion may thus stand as a task on its own or serve as a component task within MDS or other text generation settings.

However, the available resources for fusing sentences which exhibit significant content overlap are still lacking, with the most recent datasets containing only several hundreds of examples \cite{mckeown-2010,thadani-mckeown-2013-supervised}, impeding further research.
In this work, we follow \citet{thadani-mckeown-2013-supervised} and extend their described sentence fusion dataset, which is derived from Pyramid annotations created for MDS evaluation \cite{nenkova-passonneau-2004-pyramid}. Table \ref{tab:scu_ex} illustrates a single fusion instance taken from the \citet{thadani-mckeown-2013-supervised} dataset, where sentences (a-d), whose contents overlap, are taken from different reference summaries for the same source texts. The overlapping content parts were annotated by experts into SCU (Summary Content Unit) spans (in bold) and then summarized into a short sentence, denoted as the gold fusion label.

\begin{table}[t!]
\centering
\resizebox{\columnwidth}{!}{%
\begin{tabular}{@{}p{0.06\columnwidth}p{0.95\columnwidth}@{}}
\textbf{\footnotesize{a.}} & \footnotesize{\textbf{Fisheries in parts of the Philippines have been decimated} \textbf{by the use of cyanide}} in fishing.\\
\textbf{\footnotesize{b.}} &\footnotesize{Philippine fishermen use \textbf{cyanide in fishing}, needlessly \textbf{destroying immature fish}.}\\
\textbf{\footnotesize{c.}}&\footnotesize{Sodium \textbf{cyanide use by fisherman decimates fish}.}\\
\textbf{\footnotesize{d.}} &\footnotesize{In the Philippines some fishermen use homemade explosives and \textbf{cyanide for driving fish away} \textbf{from reefs and into nets}.} \\ \hline
\textbf{\footnotesize{Label }}&\footnotesize{ Sodium cyanide use by fisherman decimates fish}
\end{tabular}%
}
\caption{Sentence fusion example. (a-d) are the input sentences, originating from different expert summaries. Spans that are considered as contributing to the same single unit of content (SCU) are in bold. Label represents the gold SCU label, representing the fusion output.
}
\label{tab:scu_ex}
\end{table}

We find that the heuristics and filters applied on their original dataset result in short and highly related sentences, which may not reflect more complex and long sentences that are often found in multi-text consolidation tasks.
Moreover, their dataset uses exclusively sentences from expert summaries, which tend to be short and concise, and exclude the actual source documents that are used in practice for summarization.
The resulting high similarity within input sentences makes them amenable to extractive methods, where a representative sentence can be selected as the summary, curbing the efforts to develop an abstractive fusion of sentences. 
In this paper, we remove most of their filters, relabel a portion of the instances, and supplement the data with sentences taken from source documents, 
while providing more challenging examples that better reflect realistic multi-source summarization tasks.

Our contribution therefore is an extended sentence-fusion dataset\footnote{Our Code and data can be found here: \url{https://github.com/DanielaBWeiss/Extending-Sentence-Fusion-Resources}}, more than 3x times larger than its original, with examples extracted from both summary and document sources, along with a manual relabeling of 18\% of our target labels to better reflect the information overlap. We show that our final extended dataset is more representative of related and overlapping sentences from multiple sources in the wild.
In addition, we provide fusion baseline models trained on both the original and our extended datasets. We show that a model trained on our extended datasets outperforms in ROUGE metrics \citep{lin-2004-rouge} on the original test set, while also seeming to generate better outputs that reflect the true content intersection of the inputs. Given that sentence fusion was originally introduced as a step in a multi-document summary pipeline \cite{barzilay-mckeown-2005, marsi-krahmer-2005-explorations,mckeown-2010,thadani-mckeown-2013-supervised}, we believe that progress on this focused task may lead to new insights and breakthroughs in the larger scoped multi-document summarization setting, as well as additional text consolidation tasks.

\section{Related Work}
\label{background}
Sentence fusion is a sentence intersection generation task introduced in the context of multi-documents. The task takes as input multiple sentences that share overlapping content, and produces a single sentence focused on that overlap \cite{barzilay-mckeown-2005, filippova-strube-2008-sentence, marsi-krahmer-2005-explorations,mckeown-2010,thadani-mckeown-2013-supervised}. Several other variants of sentence fusion have also been explored, such as sentence union (fusing the union of information in the input) \cite{marsi-krahmer-2005-explorations}, and even strict sentence intersection \cite{thadani-mckeown-2011-towards} (producing strictly the intersection and nothing more). For multi-text summarization however, a ``looser'' sense of sentence intersection is desired, since redundant content is most likely salient, yet additional non-overlapping information may still be relevant for a final summarized sentence. For this reason, our extended dataset follows this ``looser'' sense of sentence intersection as was used and described in \citet{barzilay-mckeown-2005}, \citet{mckeown-2010} and \citet{thadani-mckeown-2013-supervised}.

It should be noted that, in recent years, the notion of \textit{sentence fusion} been used also to denote a quite different task variant, called ``disparate'' sentence fusion \cite{elsner-santhanam-2011-learning,geva2019discofuse,lebanoff-etal-2019-scoring,lebanoff-etal-2020-poc}. Under this variant, the fused sentences do not exhibit considerable content overlap but are rather related in discourse. 
Accordingly, this task variant is concerned with ``gluing" the input sentences into a longer sentence, while generating appropriate discourse structure, possibly inserting certain discourse connectives. 
This task is fundamentally different than the task we address, of fusing significantly redundant information, which is particularly relevant in the context of consolidating information from \textit{multiple} documents (while discourse-oriented fusion is more often considered in the context of single-document summarization).

\begin{table}[t!]
\centering
\resizebox{\columnwidth}{!}{%
\begin{tabular}{@{}p{0.05\columnwidth}@{}p{0.15\columnwidth}@{}p{0.8\columnwidth}@{}}
 & \textbf{Filter} & \textbf{Filtered SCU Labels} \\ \hline
1 & \footnotesize{SL} & Saudi Arabia urged withdrawal \\
2 & \footnotesize{NV} & Resignation of Prime Minister Karami and his government \\
3 & \footnotesize{NV | SL} & Murder in Boulder, Colorado \\
4 & \footnotesize{NV} & Confirmed bird flu cases in Hong Kong  \\
\\
& & \textbf{Filtered SCU Instance} \\
5 & \multicolumn{2}{p{0.95\columnwidth}}{Statoil admitted responsibility for the leak}\\ \hline
A & \multicolumn{2}{p{0.95\columnwidth}}{Statoil's internal investigation acknowledged inadequate planning and a lack of risk appreciation led to the leak. }\\
B & \multicolumn{2}{p{0.95\columnwidth}}{ Statoil admitted the leak resulted from inadequate planning and appreciation of risk, and failure to observe governing documentation.}
\end{tabular}%
}
\caption{Originally filtered SCU instances in \textsc{PyrFus} \citep{thadani-mckeown-2013-supervised}. \textsc{NV} -- No Verb, \textsc{SL} -- Short Label. Examples 1-4 are SCU labels that were automatically filtered based on the label alone, while the last example was filtered due to the SCU contributing spans in A and B being much longer than the label itself, expressed as a short abstract paraphrase.}
\label{tab:filtered_ex}
\end{table}
\section{Data Collection}
The original established dataset \citep{thadani-mckeown-2013-supervised} for sentence fusion leverages annotations made during post-hoc evaluation of multi-document summarization systems.
It is imperative to inspect the origin of the data and how it was re-purposed for this task (\S\ref{orig_filters}), as well as review the different processing steps implemented in previous works (\S\ref{sec:preprocess}), in order to assess our revisions and supplements (\S\ref{new_filters}).

\subsection{From Summary Evaluation to Sentence Fusion}
\label{orig_filters}
The pyramid method, proposed by \citet{nenkova-passonneau-2004-pyramid}, is a well-known  evaluation method for content selection in summarization, which was used extensively in the DUC\footnote{https://www-nlpir.nist.gov/projects/duc/data.html, years used 2005-2008} and TAC\footnote{https://tac.nist.gov/, years used 2009-2011} benchmarks for MDS.
% Two NIST organized conferences, the Document Understanding Conference (DUC) and the Text Analysis Conference (TAC) had established an expert evaluation procedure for multi-document summarization. 
Applying this method, informative content in different documents is dissected into small information units, and equivalent units are aggregated across documents.
The resulting clusters of largely-equivalent information units, each centered around a single statement, were utilized in subsequent work to form the basis for sentence fusion examples  \cite{thadani-mckeown-2013-supervised}. 

In DUC and TAC, a set of related documents over the same topic was summarized by at least four expert annotators, producing four or more short \emph{reference} summaries per topic.
Then, the \emph{reference} summaries were further analyzed and divided into a set of informational units named Summary Content Units (SCUs). Each SCU denotes a single short statement, e.g. \emph{cyanide use by fisherman decimates fish} (see bold spans in \autoref{tab:scu_ex}), and may be expressed in multiple summaries (and source documents) under different manifestations. 

To mark an SCU, the annotator marks spans of text that directly contribute to it (SCU contributors), and labels the SCU by writing a concise statement in natural language, which is called SCU Label. 
The same label is applied to equivalent content-units across documents, grouping together span contributions from different sources. 
\autoref{tab:scu_ex} presents an example of an SCU cluster, containing four sentences with contributing spans (in bold), along with their associated SCU label.
When evaluating an MDS system, a generated summary is scored proportionally to the number of SCUs whose information is covered by the summary, where each SCU is weighted its frequency in the gold reference summaries.
To create a sample for sentence fusion, the sentences in each SCU cluster are taken as input, and the SCU label, which concisely summarizes the contributing spans in the cluster, is considered as the gold text for the targeted fusion output. 
% Next, we will review in detail the different processing steps applied to the evaluation data to form a fusion dataset.

\subsection{Pre-processing for Generating Fusion Examples}
\label{sec:preprocess}
\citet{thadani-mckeown-2013-supervised} applied several pre-processing steps to generate a fusion dataset (termed here as \textsc{PyrFus}) from SCUs. While the original intention was to reduce noisy samples, these steps also removed a significant portion of challenging fusion instances. 
The applied steps include \emph{discarding} all clusters that: 
%\ido{Have consistent punctuation between items. Since colon is not a sentence separator, best end each item with a semi colon, and start the item without capitalizing the first word} 
(1) have more than 4 contributing sentences; (2) have SCU labels that don't contain a verb after the first token; 
(3) have SCU labels and source sentences with less than 5 words or more than 100; (4) have contributing spans that are shorter than half of their source sentence; (5) have SCU labels that are shorter than half of the shortest contributing span in the input; (6) have SCU labels with non-shared tokens in any of the source sentences.

Filters (2) and (3) were used because not all SCU labels in the original data were grammatical sentences, while (4) and (5) were applied to remove inputs with too little mutual content, and often the SCU label did not cover the whole information intersection of the input sentences.\footnote{Since SCUs mostly cover single propositions, sometimes the same cluster of summary sentences will share multiple SCU labels, but are split to individual fusion instances given the annotation protocol of the pyramid data.}
Filter (6) was applied so that all gold fusion labels would be lexically reachable without the need for paraphrasing.
The resulting dataset was the largest 
available source for supervised sentence fusion focused on multi-text, until this work, with a total of 1705 fusion instances \footnote{The number originally reported is slightly higher, while this is the number we were able to reproduce using the author's published code.}, compared to the previously available fusion data of 300 instances \cite{mckeown-2010}. 

\subsection{Extending Pyramid Fusion}
\label{new_filters}
While \citet{thadani-mckeown-2013-supervised} (\textsc{PyrFus}) created a polished dataset, we argue that the skipped clusters would more closely resemble document sentences used in the multi-document summarization task, and learning from such examples is feasible given contemporary models.
Additionally, DUC also made available the SCU Marked Corpus \cite{scu-marked-corpus}, which automatically maps \emph{source} document sentences to SCU labels using text match. We make use of these mappings to extend our dataset with document sentences, which were overlooked in \textsc{PyrFus}.

We found that certain filters were safe to remove while retaining a successful representation of the sentence intersection task. For instance, the majority of SCU labels that do not have a verb use verb nominalizations, or that sentences beginning with a verb are coherent.
Table \ref{tab:filtered_ex} shows a few examples of fusion instances originally filtered in \textsc{PyrFus} that we have kept in our extended version of this dataset, termed \textsc{PyrFus++}.
Similarly, the majority of SCU labels of length 4 were also found to be coherent and descriptive of their input sentences, and therefore we decided to discard SCU clusters only if they do not meet this threshold. Shorter SCU labels (length 3 and below) were deemed too short to describe a complete summarized sentence (see examples in Appendix \S \ref{filtered_examples}).
%\pl{Put examples here.}
Additionally, we allow SCU clusters that have low overlap between their label and their marked contributing spans (see ex. 5 in Table \ref{tab:filtered_ex}). 
And finally, we keep fusion instances whose SCU label tokens are not fully covered by their input sentences to allow for paraphrastic examples.

Once we removed most of the filtering pipeline of \textsc{PyrFus}, we noticed that almost 20\% of the fusion input clusters share more than one SCU label (i.e. they share more than a single informative proposition). To accommodate the target label of such instances, we manually re-label such clusters using all shared SCU labels into a single sentence. For example, for the following two SCU labels: \textit{Clinical trials typically involve three phases} and \textit{Clinical trials involve an average of 200 patients per trial}, a new merged fusion label would be:  \textit{Clinical trials typically involve three phases and an average of 200 patients per trial}.

In total, we have extended the fusion dataset from its original 1705 instances to 7502, creating a varied dataset using both reference summary and document sentences.
We suggest that the additional instances previously skipped, along with the modifications we introduced, would more closely resemble real-world  content fusion challenges in multi-document summarization and related settings. 

\begin{table}[t!]
\centering
\resizebox{\columnwidth}{!}{%
\begin{tabular}{llllll}
\textbf{Fusion Data} & \textbf{Total} & \textbf{Avg Clus.} & \textbf{$\mathbf{R_1}$ L-to-S} & \textbf{$\mathbf{R_1}$ S-to-S} \\
\citet{lebanoff-etal-2020-poc} & \textbf{1599} & 2 & 32.7 & 15.0 \\
\textsc{PyrFus} & \textbf{1705} & 2.8 & 46.5 & 35.0 \\
\textsc{$\Delta$-PyrFus} & \textbf{5842} & 3.3 & 34.6 & 31.6 \\
\textsc{PyrFus++} & \textbf{7502} & 3.3 & 37.8  & 32.2
\end{tabular}%
}
\caption{Comparisons of different fusion datasets and variations. \citet{lebanoff-etal-2020-poc} introduced a disparate-fusion dataset, containing exactly 2 input sentences from a single document. L-to-S and S-to-S refer to label-to-sentence and sentence-to-sentence ROUGE scores, respectively.}
\label{tab:fus-stats}
\end{table}
\section{Data Analysis}
\label{data_stats}
We would like to assess the similarity of the source sentences among themselves and also against their fusion target. To that end, we use the ROUGE \cite{lin-2004-rouge} metric as a proxy to similarity and calculate the micro average of ROUGE $R_1$ (measuring word overlap), between every input sentence to its SCU label, and between every pair of sentences in the same cluster.
\autoref{tab:fus-stats} compares a few established fusion datasets: The ``disparate'' fusion corpus \citep{lebanoff-etal-2020-poc} with (mostly non-overlapping) sentences originating from a single document (see \S \ref{background}), i.e. not summary sentences; \textsc{PyrFus} represents the work presented in \citet{thadani-mckeown-2013-supervised}, \textsc{PyrFus++} is our revised and extended dataset presented in this work, and $\Delta$-PyrFus refers to the part of our dataset containing SCU clusters that were overlooked in \textsc{PyrFus}.

We first note that the word overlap among source sentences ($R_1$ S-to-S) is much lower for the disparate fusion set, as expected by the nature of this dataset.
This is in contrast to the datasets originating from multi-document sources, indicating that in the former data, the input sentences to fuse share less information. This reinforces our claim that in a true multi-document setting a system will be challenged with dealing with significantly more redundant information, and this has to be applied specifically addressed.

Unsurprisingly, \textsc{PyrFus} contains a much higher source-to-target word overlap ($R_1$ L-to-S), given that the applied pre-processing explicitly removed instances with less overlap between the SCU label and the source sentences. 
Adding those instances back to \textsc{PyrFus++} lowered this metric, making the task more challenging, and realistic.
Notably, the difference in ($R_1$ S-to-S) sentence-to-sentence similarity between the original \textsc{PyrFus} and the added portion  \textsc{$\Delta$-PyrFus} is only of 3.4 rouge points. This should indicate that the input sentences remain highly related, even after removing the pre-processing pipeline applied originally, making them viable fusion instances. 

Overall we note that while some noisy instances were introduced into \textsc{PyrFus++}, the new fusion clusters express high relatedness between the source sentences and their label. This is also suggested by the level of sentence-to-label word-overlap in \textsc{$\Delta$-PyrFus} being roughly the same as the sentence-to-sentence word overlap in the original \textsc{PyrFus}.  
\section{Baselines}
\begin{table}[t!]
\centering
\begin{tabular}{llll}
\textbf{Train Data} & \textbf{Dev} & \textbf{Test} & \textbf{Test++} \\
\textsc{PyrFus} & 36.4 & 40.9 & -- \\
\textsc{PyrFus++} & \textbf{41.5} & \textbf{46.9} & \textbf{32.7}
\end{tabular}%
% }
\caption{Rouge-2 F1 results for the baseline model (BART). Rows are training sets and columns are evaluation sets. Test++ refers to the test set of the extended \textsc{PyrFus++} dataset. The other evaluation splits refer to the original \textsc{PyrFus} data.
}
\label{tab:baseline}
\end{table}
We implement a newer modern baseline for \textsc{PyrFus} \citep{thadani-mckeown-2013-supervised}, which outperforms their pre-neural one\footnote{\text{PyrFus} evaluation used bigram-F1 \citep{unno-etal-2006-trimming} that is similar to Rouge-2 F1, reporting 24.92 points for their best model. We use the widely accepted Rouge metric to be inline with contemporary works.}.
To that end, we employ the pre-trained auto-encoder BART \cite{lewis2020bart} as our end-to-end generation model due to its demonstrated performance on summarization tasks.

Results, shown in \autoref{tab:baseline}, were measured with the Rouge-2 F1 metric on the original \textsc{PyrFus} evaluation splits. They clearly show that a fusion model trained on our extended data (\textsc{PyrFus++}) significantly outperforms the same model trained on the original training data, by roughly 6 $R_2$ points. Notably, the model trained on \textsc{PyrFus++} scored 14 points lower on its own test set, indicating that the new dataset is much more challenging, and yet enables the model to reach better generalizations.

Examining the outputs of both models we find that most fusion outputs are similar and are often extracted from source sentences \footnote{This characterizes both training sets, since the original Pyramid data contains many extractive SCU labels \citep{thadani-mckeown-2013-supervised}}. Yet, we also notice that the model trained on \textsc{PyrFus++}  tends to select more salient and shared content from the input. For example, using the same input sentences as in \autoref{tab:scu_ex}, the produced labels of both models are shown in \autoref{tab:baseline_ex}, where both are lexically similar to the source sentences and to the gold SCU label. However, the model trained on \textsc{PyrFus} does not include a critical detail that all input sentences discuss -- fish decimation, while the \textsc{PyrFus++}-trained model correctly includes it. Such instances show the necessity of a large and realistic fusion dataset for model training.
\begin{table}[t!]
\centering
\resizebox{\columnwidth}{!}{%
\begin{tabular}{@{}p{0.2\columnwidth}p{0.8\columnwidth}@{}}
\textbf{\footnotesize{SCU Label}}&\footnotesize{Sodium cyanide use by fisherman \textbf{decimates} fish}\\ \hline
\textbf{\footnotesize{\textsc{PyrFus}}}&\footnotesize{In the Philippines some fishermen use cyanide in fishing}\\
\textbf{\footnotesize{\textsc{PyrFus++}}}& \footnotesize{In the Philippines cyanide use by fisherman \textbf{decimates} fish}
\end{tabular}%
}
\caption{
The gold SCU label vs the predictions made by the baseline model trained on \textsc{PyrFus} and \textsc{PyrFus++}
}
\label{tab:baseline_ex}
\end{table}

\section{Conclusion}
In this work we extended a sentence fusion dataset by more than 3 times its original size, while relabeling some of the data. The new dataset includes more complex and relevant training instances, better reflecting those that could be found in ``the wild'', and thus facilitates further research on data consolidation in multi-text tasks. In addition, we train baseline fusion models and show that when trained on our extended data we achieve notably better performance on the original available fusion test set, while also generating qualitatively better (``loose") sentence intersections.

\bibliography{tacl2018}
\bibliographystyle{acl_natbib}

\appendix
\section{Examples of Filtered SCU Labels}
\label{filtered_examples}
\begin{table}[ht!]
\centering
% \resizebox{\columnwidth}{!}{%
\begin{tabular}{ll}
 & \textbf{Filtered SCU Labels of Length 3} \\
1 & \textit{Water being diverted} \\
2 & \textit{FARC commits slaughters} \\
3 & \textit{There were floods} \\
4 & \textit{Adverse reactions reported}
\end{tabular}%
% }
\caption{Examples of SCU Labels of length 3 which were not included in this or previous works.}
\label{tab:appendix_filtered_examples}
\end{table}

\section{Extending Pyramid-based Fusion Data}
For the fusion instances containing summary source sentences as fusion inputs, we used the same years reported in \citet{thadani-mckeown-2013-supervised} (years 2005-2007 for DUC and 2008-2011 for TAC). The source document sentences found in \citet{scu-marked-corpus} were made available from 2005-2008. We made use of all the years except 2005, since we found this year to be containing more varied documents within a topic, which yielded noisier automatic alignments between SCU labels and source document sentences.
\end{document}